\documentclass[letterpaper, 10 pt, conference]{ieeeconf}  
\usepackage{FG2020}
\usepackage{gensymb}
\usepackage{times}
\usepackage{epsfig}
\usepackage{graphicx}
\usepackage{amsmath}
\usepackage{amssymb}
\usepackage{bm}
\usepackage{fixltx2e}
\usepackage{placeins}
\usepackage{gensymb}
\usepackage{tablefootnote}
\usepackage{tabularx,multirow,booktabs,blindtext}
\usepackage[multiple]{footmisc}
\usepackage[font=footnotesize]{caption}
\usepackage[colorlinks,linkcolor = red]{hyperref}
\usepackage{textcomp}
\usepackage{lipsum}

\FGfinalcopy 

\IEEEoverridecommandlockouts                              
\title{\LARGE \bf
In-the-wild Drowsiness Detection from Facial Expressions
}

\author{\parbox{16cm}{\centering
    {\large Ajjen Joshi, Survi Kyal, Sandipan Banerjee, Taniya Mishra}\\
    {\normalsize
    Affectiva, Boston, MA, USA\\
    \{ajjen.joshi, survi.kyal, sandipan.banerjee, taniya.mishra\}@affectiva.com}}
}

\begin{document}

\ifFGfinal
\thispagestyle{empty}
\pagestyle{empty}
\else
\author{Anonymous HSIM20 submission}
\pagestyle{plain}
\fi
\maketitle

\begin{abstract}

Driving in a state of drowsiness is a major cause of road accidents, resulting in tremendous damage to life and property. Developing robust, automatic, real-time systems that can infer drowsiness states of drivers has the potential of making life-saving impact. However, developing drowsiness detection systems that work well in real-world scenarios is challenging because of the difficulties associated with collecting high-volume realistic drowsy data and modeling the complex temporal dynamics of evolving drowsy states. In this paper, we propose a data collection protocol that involves outfitting vehicles of overnight shift workers with camera kits that record their faces while driving. We develop a drowsiness annotation guideline to enable humans to label the collected videos into 4 levels of drowsiness: `alert', `slightly drowsy', `moderately drowsy' and `extremely drowsy'. We experiment with different convolutional and temporal neural network architectures to predict drowsiness states from pose, expression and emotion-based representation of the input video of the driver's face. Our best performing model achieves a macro ROC-AUC of 0.78, compared to 0.72 for a baseline model.  

\end{abstract}

\section{INTRODUCTION}

Drowsy driving is dangerous, causing accidents that kill or injure thousands of people every year. It is estimated that between 2.3\% to 2.5\% of all police-reported fatal crashes between 2011-2015 involved drowsy driving, resulting in more than 4,000 deaths \cite{Crash_Stats}. Equipping vehicles with the ability to detect signs of drowsiness in drivers can potentially save lives. However, developing robust drowsiness detection methods continues to be a challenging research problem.   

\begin{figure}[t]
      \centering
      \includegraphics[width=0.85\linewidth]{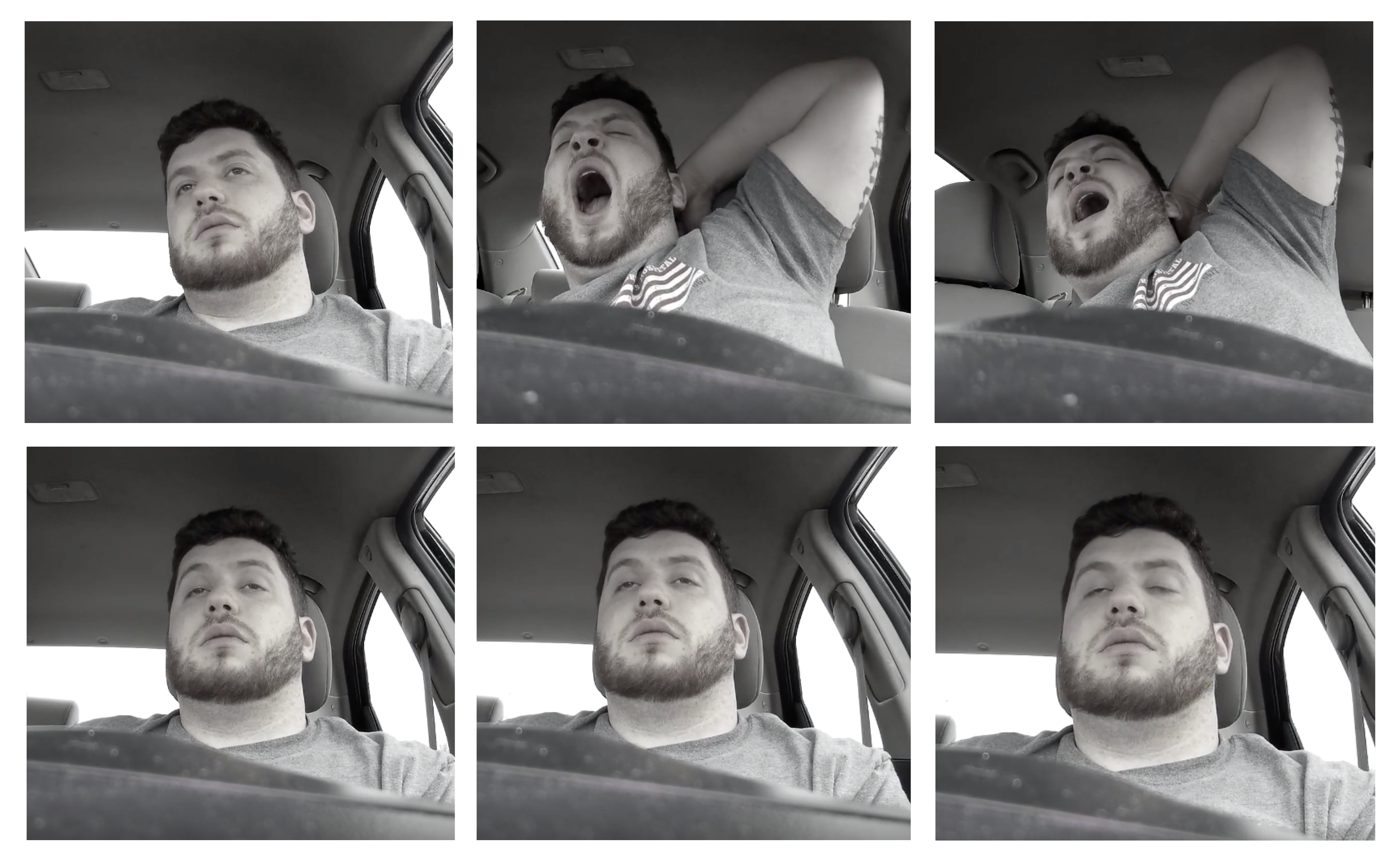}
      \caption{Sample frames from a video where the driver transitions from `Alert' to `Slightly Drowsy' (Top left) to `Moderately/Extremely Drowsy' (Bottom Right). We present a system that can classify drowsy states of real-world drivers based on video sequences captured by an in-cabin camera.}
      \label{fig:Figure1}
\vspace{-0.1cm}
\end{figure}

A variety of methods have been used to detect drowsiness. Most technologies deployed in the automotive industry rely on vehicular behavior, such as distance from lane markers on the road or steering behavior \cite{khan2019comprehensive}. Limitations of vehicular metrics are that they may not be uniform across all driving scenarios, varying by road or climatic conditions. Moreover, such systems are not widely in use and are often available only in select vehicles and brands. Arguably, the richest source of signal depicting drowsy behavior is the face of the driver. With recent advances in computer vision and machine learning, particularly in facial analysis, developing drowsiness detection system based on a driver-facing camera seems promising and is the focus of our work (Fig. \ref{fig:Figure1}).

There are several challenges to building real-world systems capable of accurately classifying drowsiness levels of a driver by analyzing their facial expressions. 
Drowsiness is a complex phenomenon, signs of which manifest in a wide variety of ways in different people's faces. The complexity of such behavior is therefore not conducive to models that rely on simple, rule-based classification of facial expression patterns; consequently making it a problem for which a data-driven machine-learning based approach is more appropriate. 

\begin{figure*}
  \centering
  \includegraphics[width=\textwidth]{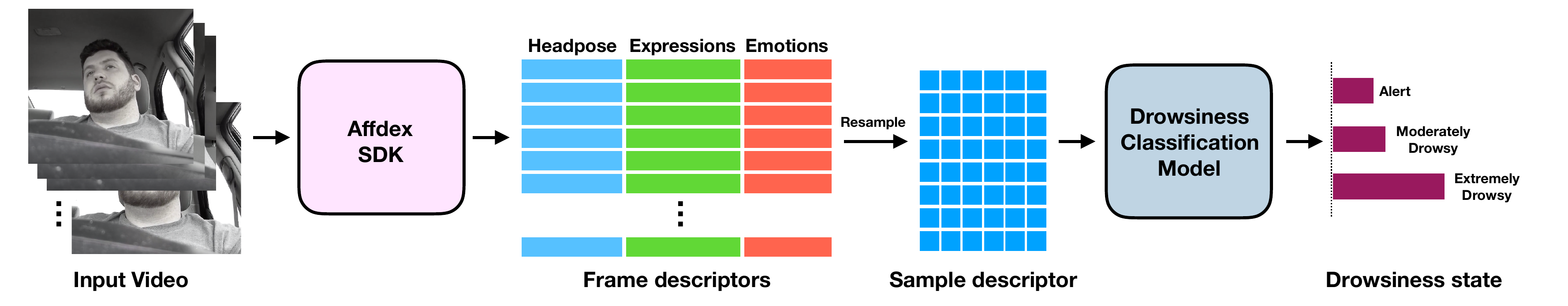}
  \caption{Schema of our drowsiness classification pipeline. Our models are trained on a representation of a 10 second video. We process each frame of a 10-second sample using the Affdex SDK \cite{Affdex_SDK}, which outputs a descriptor for every frame, consisting of estimates of head-pose, facial expressions and emotions of the driver. This sequence of descriptors is resampled into a 18$\times$100-dimensional vector, which is input to our drowsiness classification model.}
  \label{fig:Schema}
\vspace{-0.6cm}
\end{figure*}

A machine-learning based drowsiness classification model requires a large amount of training data encompassing a wide range of drowsy driving behavior. However, the safety challenges associated with drowsy driving make the collection of real-world data difficult. This is why existing drowsy datasets either involve participants acting out pre-defined sequences of drowsy behavior or are collected in the lab using elaborate protocols to elicit drowsiness. Neither methods result in the collected data being completely reflective of real-world drowsy driving behavior. Thus, models trained on lab-collected datasets struggle to generalize to in-the-wild real-world scenarios. But collecting a naturalistic, drowsy driving dataset comes with its own set of challenges. For example, in real-world driving, actual instances of drowsy behavior (as a fraction of total time spent driving) occur very infrequently. This results in needing to collect a very large driving corpus to accumulate a meaningful amount of drowsy behavior, sufficient to train a robust model. 

Another challenge associated with modeling drowsiness is the annotation of collected data. Drowsiness is a time-varying phenomenon that involves continuously transitions between drowsy states. It is not obvious what the right number of drowsy states that should be annotated is and what the defining characteristics of each state are. Moreover, the boundaries between these states can be ambiguous, i.e. it is difficult to ascertain when a certain drowsiness state begins and ends. In addition, manifestations of drowsy behavior maybe be perceived differently by different annotators, resulting in a significant percent of drowsy behaviors labeled without adequate annotator agreement.  

Our contributions in this work are threefold: first, we implemented a data collection protocol that involved setting up vehicles of overnight shift workers with camera kits that record their faces while driving. Second, we developed a drowsiness annotation guideline to label the collected videos into 4 levels of drowsiness: `alert', `slightly drowsy', `moderately drowsy' and `extremely drowsy'. Third, we experimented with various neural network topologies that map a pose, expressions and emotions-based representation of the input video of the driver's face into the correct drowsiness state. We found a 2D-CNN trained on a grid of facial features over time achieved the best performance. A schema of our classification pipeline is illustrated in Fig. \ref{fig:Schema}.

\section{RELATED WORK}
{\bf Collecting Drowsiness Data}: Rosario et al. presented a study of driver drowsiness, where  physiological signals, eye closure, pressures on the seat, and vehicular behavioral measures were recorded in a driving simulator, during a test with 20 volunteers in an environment that induced drowsiness \cite{de2010controlled}. Based on this study, a control signal that combined EEG and eye closure (PERCLOS) was proposed to classify the different drowsiness states. 

Other drowsiness-related datasets include YawDD, which is focused only on the act of yawning \cite{abtahi2011driver}; NTHU-DDD, which has 36 participants ``acting''  drowsy sequences (with yawning, slow blink rates, frequent nodding and falling asleep) and non-drowsy sequences (regular blinks, talking, laughing, looking at both sides) in a laboratory simulator \cite{weng2016driver}; and, DROZY, where 14 participants undergoing a psychomotor vigilance test (PVT) were filmed using the Microsoft Kinect, which provided both color and depth images, and their electroencephalograph (EEG), electrooculogram (EOG), electrocardiogram (ECG) and electromyogram (EMG) signals were also collected \cite{massoz2016ulg}. It can be argued that none of these datasets are reflective of real-world drowsy driving behavior because they are either acted (\cite{weng2016driver}, \cite{abtahi2011driver}) or collected in simulators (\cite{massoz2016ulg}, \cite{de2010controlled}) that behave quite differently than real vehicles.

{\bf Measuring Drowsiness}: The presence and intensity of drowsy behavior has been annotated in a variety of ways. A popular method is to use self-reports such as KSS \cite{aakerstedt1990subjective} with its 9 levels of drowsiness, which is administered before, after and/or during the driving task. Other drowsiness/sleepiness self-reported scales include the Standford Sleepiness Scale (SSS) \cite{hoddes1972stanford} and the Epsworth Sleepiness Scales (ESS) \cite{johns1991new}.

It is often impractical for drivers to provide intermittent self-reports during a driving task. The Observer Rating of Drowsiness (ORD) is a scaled ratings protocol that external observers utilize to rate drowsiness in videos \cite{wierwille1994evaluation}. ORD has 5 levels of drowsiness (`Not drowsy', `Slightly drowsy', `Moderately drowsy', `Very drowsy' and `Extremely drowsy'). Observers look for behavioral indicators of drowsiness such as eyelid closures, yawns, a vacant stare, body movements or the head falling backward or forward. In practice, however, it is challenging to annotate drowsy behavior into 5 different levels with a sufficient degree of annotator agreement.

{\bf Modeling Drowsiness}: A number of salient facial behaviors are associated with drowsiness, such as rapid and constant blinking, nodding or swinging of the head, and frequent yawning \cite{fan2009yawning}. Most studies on using behavioral approaches to determine drowsiness, focus on eye behavior \cite{bergasa2006real}. PERCLOS, the percentage of time in a minute the eyes are at least 80\% closed, has been analyzed in many studies \cite{dinges1998perclos,mckinley2011evaluation} and used as a reliable proxy for drowsiness. Different from the rule-based PERCLOS, machine learning models that focus on eye behavior have also been developed \cite{mandal2016towards, massoz2016ulg}. Eye-based drowsiness models can have limited utility because it is challenging to accurately track the eyes, for example if the driver is wearing glasses, or to learn models that generalize to all shapes and sizes of eyes.

In addition to eye state, methods to recognize driver drowsiness and fatigue states have utilized a variety of signals collected from different sensors to detect their presence and assess their intensities. Researchers have devised systems to detect these states, for example, by analyzing facial expressions \cite{dwivedi2014drowsy}, speech \cite{pir2017automatic}, posture \cite{teyeb2016towards}, physiological signals such as EEG \cite{borghini2012assessment}, and vehicle measurements that capture driving behavior \cite{thiffault2003monotony, ingre2006subjective}. Reddy et al. presented a multi-stream CNN architecture for the task of drowsiness prediction, focusing on model compression to meet the requirements of real-time inference in embedded systems \cite{reddy2017real}. Weng et al. introduced a drowsiness dataset and proposed a method based on facial feature extraction using Deep Belief Networks and temporal modeling using Hidden Markov Models \cite{weng2016driver}. Improvements on this work have been presented using various deep learning architectures, most of which rely on end-to-end training of CNN-based feature extractors and LSTMs to model temporal dynamics \cite{shih2016mstn, guo2019driver}. A survey of driver fatigue detection methods can be found in \cite{sahayadhas2012detecting, shi2017review}.

\begin{figure*}
  \centering
  \includegraphics[width=0.9\linewidth]{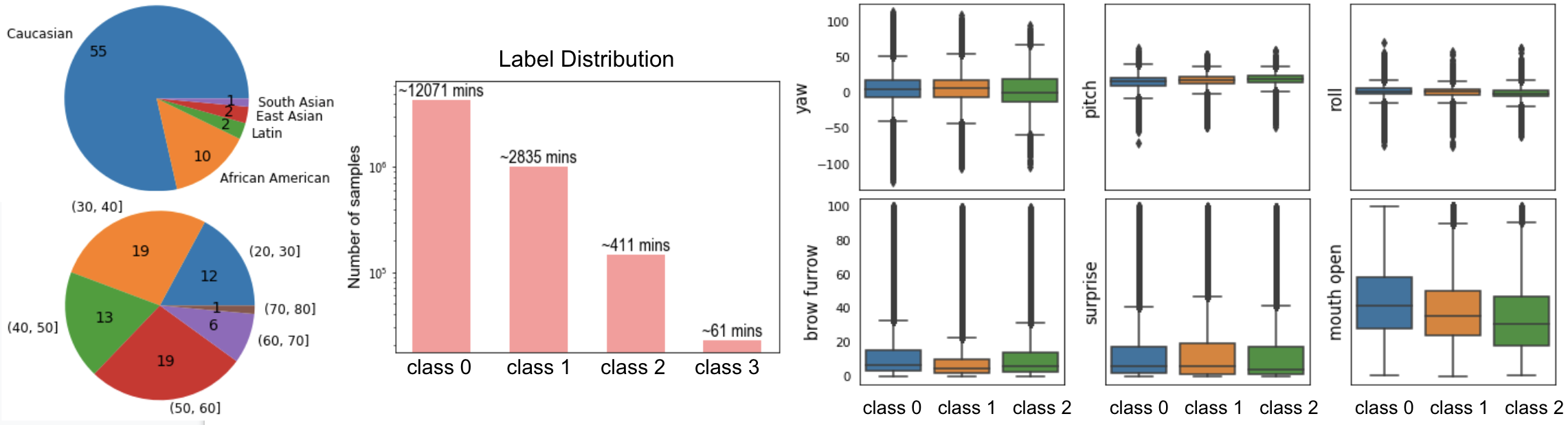}
  \caption{(Left) Pie charts illustrating the distribution of participants in the dataset in terms of ethnicity (top) and age brackets (bottom). (Middle) Plots representing the distribution of labeled classes in the dataset (Class 0 - Alert, Class 1 - Slightly Drowsy, Class 2 - Moderately Drowsy, Class 3 - Extremely Drowsy). Only 61 minutes of the dataset was labeled extremely drowsy driving, a very small fraction of the entire dataset. (Right) Box plots showing the distributions of 6 different features across different classes (where Classes 2 and 3 are merged to form Class 2 - Moderately/Extremely Drowsy).}
  \label{fig:DataAnalysis}
 \vspace{-0.6cm}
\end{figure*}

\section{Data Collection and Annotation}


The purpose of this collection was to accumulate examples of real-world drowsy driving behavior because existing datasets are either acted \cite{weng2016driver}, collected in a simulator \cite{de2010controlled, massoz2016ulg}, or consist only of a specific behavior associated with drowsiness e.g. yawning \cite{abtahi2011driver}. In order to maximize chances of capturing drowsy behavior, camera kits were installed in the vehicles of night shift workers, as they are more prone to driving in a fatigued state \cite{caruso2014negative}. 

A total of 223 participants took part in this study. The participants were recruited in Boston, Chicago and Cairo. Their vehicles were installed with cameras and they were instructed to record their faces while driving for up to two weeks. They were also instructed to keep logs of how drowsy they were before and after each drive using the KSS scale. The camera used for the collection was the Transcend Drivepro 520, which has an F/1.8 aperture, 130 degree viewing angle, records at 1080P and at 30 fps. The data was captured on a local chip storage device, which was mailed back to the study administrator along with KSS logs. The participants consented to the use of their data for research and development and were compensated \$200 for their efforts.

Videos from the driving sessions for all participants were split into 5 minute segments and associated with the KSS scores logged at the end of the drive. However, participants only reported KSS scores for less than 50\% of the driving sessions. Propagating a single KSS score reported at the end of a long drive to be the ground truth label for the entire 5-minute video would be inaccurate. However, driving sessions with high KSS scores reported at the end are more likely to contain drowsy behavior; under this assumption, we only selected videos with KSS scores reported to be 6 or higher in order to maximize the chances of selecting drowsy driving behavior and to best utilize limited annotation resources. This resulted in a dataset of 2034 5-minute videos.

\begin{table}[t]
    \centering
    \begin{tabularx}{\linewidth} {lX}
    \toprule
    Drowsy Label & Indicators\\
    \midrule
    \multirow{5}{.2\linewidth}{Alert} 
    & * fully alert individual * no indicators of drowsiness * attentive and engaged in driving * normal headpose, eyelid droop, blinks etc. \\
    \midrule
    \multirow{5}{.2\linewidth}{Slightly Drowsy}
    & * clear signs of drowsiness but otherwise alert and attentive in driving * one or more indicators of drowsiness e.g. increased blink rate and/or repeated yawning * despite being drowsy, active in operating the vehicle: looking around, engaging with passengers etc. \\
    \midrule
    \multirow{5}{.2\linewidth}{Moderately Drowsy}
    & * signs of reduced alertness due to fatigue * actively engaged in driving, but at a reduced capacity * e.g. not sufficiently alert to anticipate sudden events * fixating on single points * reduced engagement with the in-car environment * attempts to keep themselves awake: shifting positions, rubbing eyes etc. * slow and frequent blinks * drooping eyelids \\
    \midrule
    \multirow{5}{.2\linewidth}{Extremely Drowsy}
    & * fails to operate a vehicle safely due to fatigue * resting eyes, eyes closed for extended periods, or starting to fall asleep and reawaken * e.g. rapid jerking of head upright, or rapid opening of eyelids * on the verge of falling asleep, and no longer fit to operate a vehicle safely \\
    \bottomrule
    \end{tabularx}
\caption{Summary of annotation guideline for the various drowsiness states} \label{tab:annotationguideline}
\end{table}

Next, we developed an annotation guideline to label our dataset into 4 classes: ``alert'', ``slightly drowsy'', ``moderately drowsy'' and ``extremely drowsy'', which is different than the 5 class drowsiness levels described in the ORD \cite{wierwille1994evaluation}. The annotation guideline was adapted based on feedback from observing real-world driving data, given the challenges of annotating samples into 5 different classes of drowsiness.

The annotators' objective was to rate whether or not the driver was perceived by them to be experiencing drowsiness, and how far that drowsiness had progressed in terms of their ability to operate a motor vehicle. They were asked to consider a number drowsiness indicators, such as yawning, blink duration, blink rate, eyelid droop, gaze, head movement, facial expression, environmental cues (such as time of day) and other behavioral indicators (such as movement of hands etc.). They were also provided with contextual information from the entire duration of the video to aid their judgment.

Each video was labeled by 3 annotators who were randomly selected from a larger pool. For every 5-minute video, the drowsiness states were labeled at a segment-level, where annotators denoted the start and end points of a particular state of drowsiness. The label corresponding to that particular drowsiness state was then propagated to each frame in the segment. All annotators had domain experience and training in annotating facial action units as well as emotional and cognitive states based on facial signals. Furthermore, annotated data went through an independent quality assessment to ensure labeling integrity. Additionally, we only selected samples with majority annotator agreement for training. The distinguishing behaviors for the four drowsy classes are summarized in Table \ref{tab:annotationguideline}. 

\section{Data Analysis}

We observed a wide variability in camera placement since participants themselves positioned the camera and there was also variations in the make and model of the cars. For analysis and modeling, we selected only those videos that were captured with the camera placed in the steering-column location. In order to constrain the training and test set to be within a certain range of the target camera angle (camera placed on the steering column, with a near-frontal view of the driver's face), we removed a portion of data for experiments reported in this paper.

Videos taken from other camera locations, such as rear-view mirror, proved challenging for our pre-processing module (automatic face tracking and facial expressions and emotions prediction), and was not used in analysis and experiments presented here. Filtering for camera-position resulted in a total of 1450 videos in our final dataset with 70 unique participants (42 female, 28 male). 

Fig. \ref{fig:DataAnalysis} (left) illustrates the demographic and age distributions of our overall dataset. Gender and age are fairly balanced but ethnicity distribution is heavily dominated by the Caucasian group, reflecting the population of where the data collection took place. Fig. \ref{fig:DataAnalysis} (middle) shows the distribution of frames labeled across the 4 drowsiness classes as per the annotation guideline. As expected, drivers exhibited extremely drowsy driving behavior for a tiny fraction of their driving sessions (only \texttildelow61 minutes total). In order to overcome this extreme class imbalance during modeling, we merged the Moderately and Extremely drowsy classes into a single Moderately/Extremely drowsy class. 

We used the Affdex SDK \cite{Affdex_SDK}, which consists of a suite of classifiers whose outputs we use as an intermediate feature representation for each frame of the videos in our dataset. The SDK outputs a number of estimates related to the driver's face, which includes head-pose (yaw, pitch, roll), facial expressions (blink,  brow  furrow,  brow  raise,  cheek  raise,  eye  closure, mouth  open,  nose  wrinkle,  smile,  upper  lip  raise,  valence, yawn)  and  emotions  (anger,  disgust,  joy,  surprise). 
In Fig. \ref{fig:DataAnalysis} (right), we also plotted the distributions of 6 different features across different classes. As can be observed, the distributions of these features are variable (e.g. Mouth Open), differences which may be picked by downstream classifiers to distinguish between the different drowsiness classes. Drowsiness modeling is not directly dependent on illumination conditions, because it is built on representations of head-pose, facial expressions and emotions. The models outputting these intermediate representations were trained using data with a variety of illumination conditions \cite{Affdex_SDK}.

\section{Experiments}

For evaluation of our models, data from 10 participants were held-out for testing, preserving the gender, ethnicity and age distributions of the dataset, resulting in a test set of 245 videos, each with a 5-minute duration. The remaining data was split 3:1 into train and validation sets. We trained our models considering a 10-second sequence as a single sample. For our experiments, we only considered samples with a single label (i.e. we did not consider 10-second sequences where the drowsiness state transitioned from one state to another) because the temporal boundaries of drowsiness states were ambiguous. The ground truth labels were assigned by taking the majority label provided by the human annotators. In order to alleviate class imbalance, samples from different classes were selected using different strides (75 frames for `alert' and 5 frames for the `drowsy' classes). We processed each frame of a 10-second sample using the Affdex SDK \cite{Affdex_SDK}, which generates an output of 18 features for every frame, as described in Section IV. This sequence of frame descriptors are resampled into a 18$\times$100-dimensional sample descriptor.

\subsection{Baseline Drowsiness Predictor and Feature Importance}
We investigated the relative importance of the facial features in discriminating different drowsiness classes, hypothesizing that certain features are differently distributed for the various drowsiness states. For example, the average eye closure for a sample where the driver is extremely drowsy is expected to be higher than that for a sample with a fully alert driver. We therefore summarized the distributions of each of the 18 facial features by using six statistics: mean, maximum, minimum, standard deviation, skewness and kurtosis. This results in a 108-D feature representation for each sample.

\begin{figure}[t]
      \centering
      \includegraphics[width=0.75\linewidth]{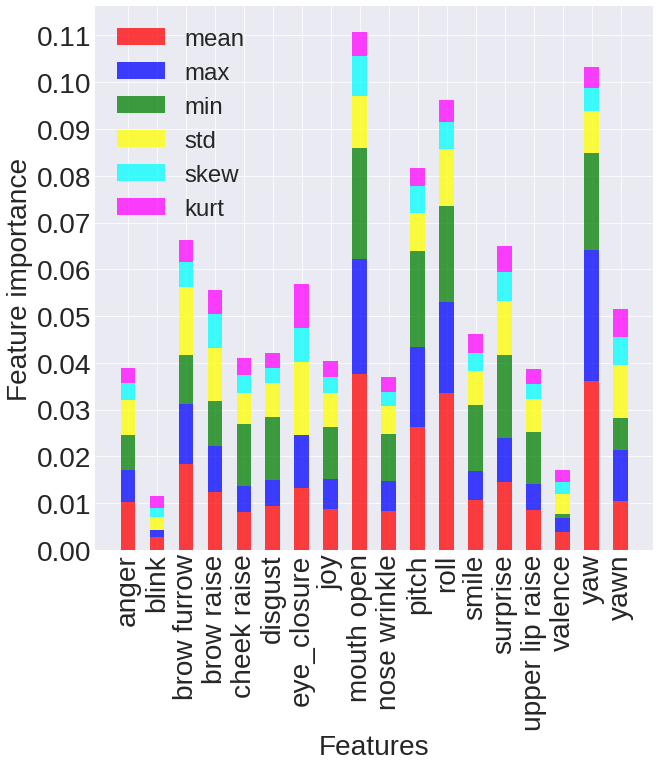}
      \caption{Feature importance computed by the baseline random forest model, aggregated for each Affdex SDK \cite{Affdex_SDK} facial metric.}
      \label{fig:featureimportance}
\end{figure}

We trained a random forest model (\textbf{RF-baseline}) (100 trees trained to a depth of 20) as our baseline drowsiness predictor. The model achieved a macro ROC-AUC of 0.72 on the test set (Table \ref{tab:results}). The importance of each SDK feature was also computed by the random forest model, aggregating the feature importance of all statistics for each facial metric. Mouth open, eye closure and head pose (yaw, pitch, roll) were found to be the most discriminative (Fig. \ref{fig:featureimportance}).

We also trained a baseline multi layer perceptron on the same 108-D statistical feature representation. The network with one hidden layer with 4 activation nodes (trained with Adam and categorical cross-entropy loss) yielded a macro ROC-AUC of 0.71 on the test set (\textbf{MLP-stats}). 

\subsection{Unsupervised Autoencoded Feature Learning}
Instead of using preset statistical scores, we trained an autoencoder model \cite{AE} to learn representations of the same length (108-D) in an unsupervised manner. By learning to map the original sample to this lower dimensional space, the model preserves only the relevant features and strips away any redundant information. Our autoencoder model comprises of an encoder network, composed of four fully connected layers, each with Leaky ReLU \cite{leakyrelu} activations, that takes in a 18$\times$100 feature vector and maps it to the 18$\times$6 space. This downsampled 18$\times$6 vector is then fed to the decoder network, an upside down version of the encoder architecture, that tries to reconstruct the original 18$\times$100 input. The loss is computed using mean absolute error between the input and reconstructed features.

Once the autoencoder was trained with our training set, we fed the trained encoder module 18$\times$100 sample descriptors from the training, validation and testing splits and extracted the generated 108-D representations. We trained the same classifier network from Sec V.B. with these lower dimensional features and generated an ROC-AUC of 0.74 on the test set (\textbf{MLP-enc}). 

\subsection{Deep Learning from Sample Descriptors}
Next, we investigated whether models trained to learn a mapping directly from the raw 18$\times$100-d sample descriptors resulted in performance better than a hand-engineered statistical feature representation or an unsupervised feature learning approach. We experimented with three neural network topologies: a) A 1-D temporal convolutional neural network with four strided 1-D convolution layers followed by dropout and global average pooling (\textbf{Conv1D-raw}), b) a 2-D convolutional neural network with three strided 2-D convolution layers followed by dropout and global average pooling (\textbf{Conv2D-raw}), and c) an LSTM network with two consecutive LSTM blocks (\textbf{LSTM-raw}). The Conv1D model learns hierarchical filters that capture discriminative temporal patterns that occur in each feature channel, whereas the Conv2D models learns a hierarchy of 2D filters that capture patterns that co-occur between the different feature channels in the sample descriptor grid. The LSTM network attempts to explicitly model the recurrent temporal dynamics unique to each drowsiness class. Training all networks for 20 epochs with Adam \cite{Adam} and categorical cross-entropy resulted in a 0.75, 0.78 and 0.77 ROC-AUC scores for (\textbf{Conv1D-raw}), (\textbf{Conv2D-raw}) and (\textbf{LSTM-raw}) respectively (Table \ref{tab:results}). 

\begin{table}[t]
\centering
\begin{tabular}{|c|c|c|c|c|c|}
\hline
\textbf{Model}  &  \textbf{AUC} & \textbf{Acc}  & \textbf{Pre} & \textbf{Rec} & \textbf{F1} \\
\hline
\hline
\textbf{RF-baseline}  & 0.72  & 0.42  & 0.39 & 0.42 & 0.39 \\
\hline
\hline
\textbf{MLP-stats}  & 0.71  & 0.58  & 0.66 & 0.58 & 0.59 \\
\hline
\textbf{MLP-raw}  & 0.71  & 0.53  & 0.64 & 0.53 & 0.53 \\
\hline
\textbf{MLP-enc}  & 0.74  & 0.50  & 0.65 & 0.50 & 0.52 \\
\hline
\hline
\textbf{Conv1D-raw}  & 0.75  & 0.59  & 0.64 & 0.59 & 0.57 \\
\hline
\textbf{Conv2D-raw}  & \textbf{0.78}  & 0.63  & \textbf{0.69} & 0.63 & \textbf{0.63} \\
\hline
\hline
\textbf{LSTM-raw}  & 0.77  & 0.57  & 0.64 & 0.57 & 0.54 \\
\hline
\hline
\textbf{Conv2D-raw + SMOTE} & 0.75 & \textbf{0.64} & \textbf{0.69} & \textbf{0.64} & \textbf{0.63} \\
\hline
\end{tabular}
\caption{Table of metrics representing the performance of our models on the test data. Here, AUC: macro-averaged AUC-ROC, Acc: Accuarcy, Pre: Weighted Precision, Rec: Weighted Recall, F1: Weighted F1 Score. Note that Acc, Pre, Rec and F1 are computed after Threshold Tuning as described in Section VI. E}
\label{tab:results}
\end{table}

\subsection{Class-specific Threshold Tuning}
The most common method for deciding the output class from the probabilities of the final layer of a model is to choose the class with the largest confidence score. However, that may not be ideal for problems where the cost of false positives is not same across all classes. In the case of a drowsiness classifier, it might be appropriate to minimize mis-classifying ``extremely drowsy'' samples at the cost of additional false positives. Therefore, we propose the following class-specific tuning. For the `Slightly Drowsy' and the `Moderately/Extremely Drowsy' classes, we independently choose thresholds that maximize \emph{True Positive Rate - (1 - False Positive Rate)}. For a test sample, if thresholds for both these classes are exceeded, we err on the side of the `Moderately/Extremely Drowsy' class. 

The difference in model predictions with and without threshold tuning is illustrated in the confusion matrices in Figure \ref{fig:conmat}, where we observe that threshold tuning results in vastly improved predictions for the `Slightly Drowsy' and `Moderately/Extremely Drowsy' classes at the cost of a slightly reduced performance for the `Alert' class.

\section{Discussion and Conclusion}
Our experimental results yielded the following interesting points of discussion.

First, from Fig. \ref{fig:featureimportance}, we see that the head pose related features (pitch, yaw and roll) are among the most predictive for drowsiness, second only to mouth-open. These results are somewhat surprising given that anecdotally it is often assumed that eye closures and yawns are the main drowsiness predictors, further bolstered by in-market drowsiness solutions that are based simply on the latter features.

The optimal performance of the Conv2D model compared to the other architectures is another interesting result, leading us to conjecture that the Conv2D model is likely learning nuanced and long-distance dependencies between the input features. One interesting future work is to learn their optimal ordering to further improve model performance. 

The improvement in accuracy of the Conv2D-raw model by augmenting it with synthetic features, generated using SMOTE \cite{chawla2002smote}, is another promising result (Table \ref{tab:results}). Adding such synthetic features to balance the training dataset pushes the model to learn more robust representations and produces a boost in almost all metrics during testing. We believe this augmentation procedure can be further enriched by introducing a GAN \cite{GAN} into the mix. Instead of generating images or video frames as traditionally done, we plan to generate synthetic samples in the feature space directly.

\begin{figure}[t]
      \centering
      \includegraphics[width=0.8\linewidth]{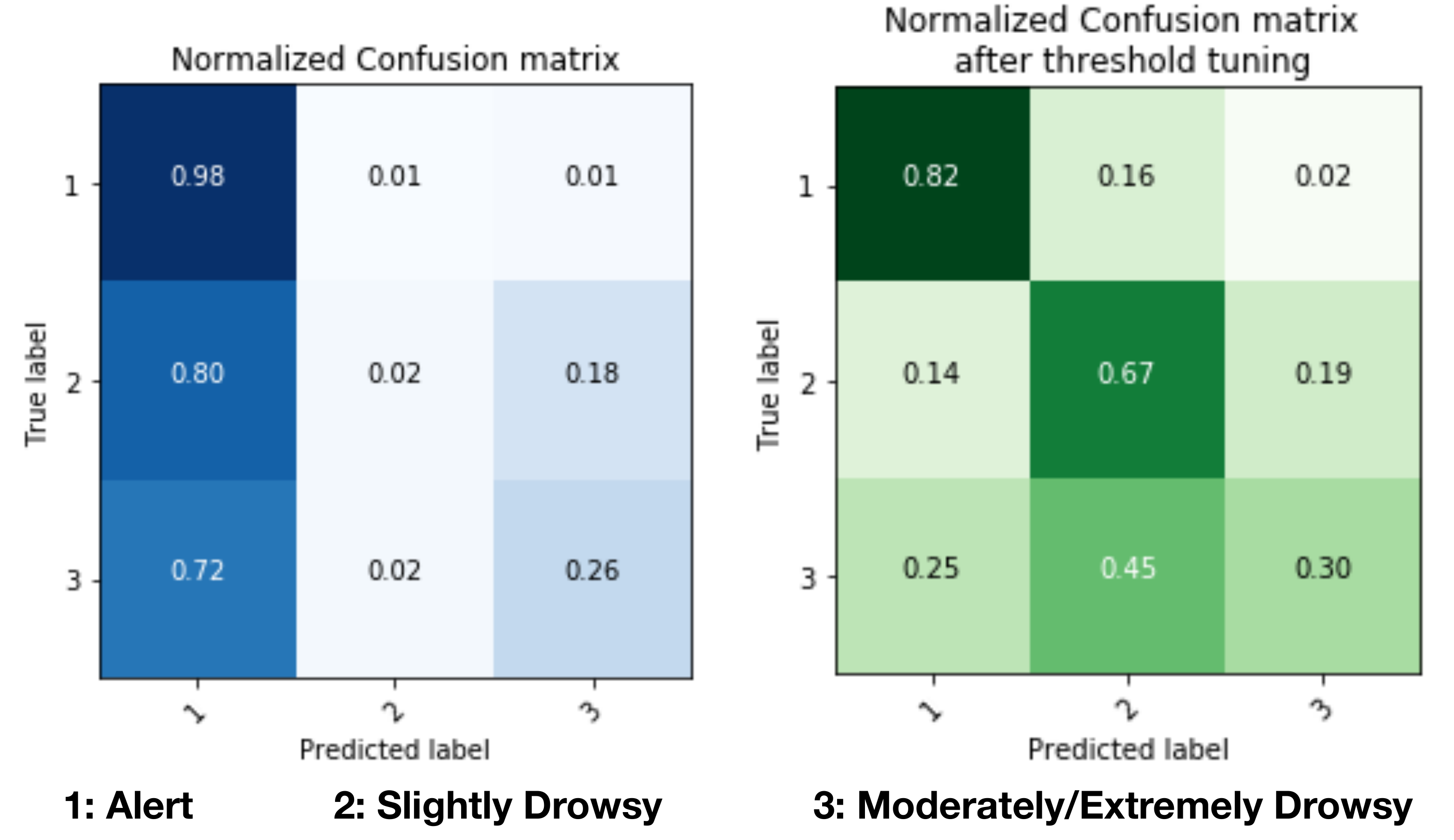}
      \caption{Confusion matrices depicting predictions of our best performing model (\textbf{Conv2D-raw}), before and after optimal thresholding.}
      \label{fig:conmat}
\vspace{-0.1cm}
\end{figure}

The substantial relative improvement in pairwise disambiguation of the different levels of drowsiness through the class-specific threshold tuning (Fig. \ref{fig:conmat}), primarily done to improve the disambiguation of the Slightly versus the Moderately/Extremely drowsy levels, underpins the importance of considering the cost of different types of misclassifications.  

On performing error analysis, we observed that the participants in the videos with most misclassifications had a non-frontal facial pose. Even though we filtered our dataset for a particular camera angle, there was still a lot of variance in how the cameras were placed in participants' cars, which resulted in decreased reliability of facial features fed to our model. Future work will involve developing models that can be personalized to different camera views. 

Deploying drowsiness detection systems that work well in real-world driving scenarios is a challenging problem. In this work, we collected a large-scale, real-world drowsy driving dataset and developed an annotation guideline to enable humans to label the collected videos into varying levels of drowsiness. We experimented with different neural network architectures to predict drowsiness  states from facial pose, expression and emotion-based temporal representations of  the input video of the driver’s face. Future work will include comparing against models that have been trained with other modalities (e.g. EEG, PERCLOS) and deploying and testing these systems in real-world scenarios.

{\small
\bibliographystyle{ieeetr}
\bibliography{references}
}

\end{document}